\documentclass[11pt,a4paper]{article}

\usepackage[utf8]{inputenc}
\usepackage[T1]{fontenc}
\usepackage{lmodern}
\usepackage[english]{babel}

\usepackage[margin=1in]{geometry}
\usepackage{setspace}
\usepackage{amsmath,amssymb}

\usepackage{graphicx}
\graphicspath{{figures/}}

\usepackage[colorlinks=true,linkcolor=blue,citecolor=blue,urlcolor=blue]{hyperref}

\usepackage{enumitem}

\usepackage[font=small,labelfont=bf]{caption}

\usepackage{authblk}

\title{Towards More Standardized AI Evaluation:\\From Models to Agents}

\author[1]{Ali El Filali$^{*}$}
\author[2]{In\`{e}s Bedar$^{*}$}
\affil[1]{G42}
\affil[2]{Independent}

\date{February 2026}

\begin{document}

\maketitle
\renewcommand{\thefootnote}{}
\footnotetext{$^{*}$Equal contribution.}
\renewcommand{\thefootnote}{\arabic{footnote}}

\begin{abstract}
Evaluation is no longer a final checkpoint in the machine learning lifecycle. As AI systems evolve from static models to compound, tool-using agents, evaluation becomes a core control function. The question is no longer ``How good is the model?'' but ``Can we trust the system to behave as intended, under change, at scale?''. Yet most evaluation practices remain anchored in assumptions inherited from the model-centric era: static benchmarks, aggregate scores, and one-off success criteria. This paper argues that such approaches are increasingly obscure rather than illuminating system behavior. We examine how evaluation pipelines themselves introduce silent failure modes, why high benchmark scores routinely mislead teams, and how agentic systems fundamentally alter the meaning of performance measurement. Rather than proposing new metrics or harder benchmarks, we aim to clarify the role of evaluation in the AI era, and especially for agents: not as performance theater, but as a measurement discipline that conditions trust, iteration, and governance in non-deterministic systems.
\end{abstract}

\section{From Output Checking to Behavior Measurement}
\label{sec:from-output}

A system is not considered usable because it works just once, but because it behaves predictably across situations, inputs, and time. We evaluate machines not because we distrust them, but because we depend on them. An aircraft is not trusted because it flies once, but because it flies repeatedly, within tolerances, across weather conditions, between different countries, etc.

At its core, AI is a tool built by humans to serve a purpose. Like any tool built and used by humans, it must be assessed against what it is supposed to do, how reliably it does it, and under which constraints. An evaluation can be defined as ``\emph{Empirical tests designed to provide early warning when a model is approaching a capability threshold. These tests are intended to trigger before a model actually reaches a dangerous capability}''~\cite{anthropic2025rsp}. Evaluation is, therefore, the discipline of building instruments that make behavior observable, comparable, and debuggable over time. In classical AI systems, evaluation was relatively contained. Models were static components: given an input, they produced an output. Evaluation focused on task-level performance using fixed datasets and aggregate metrics. Accuracy, precision, recall, or loss curves were imperfect but often sufficient proxies for quality. AI agents are not just models; they are systems that reason over multiple steps, invoke tools, interact with environments, maintain state, and adapt their behavior dynamically.

A single correct output, more than ever, proves nothing. This is why evaluation methodologies diverge sharply between classical AI and generative/agentic systems. For models, evaluation often asks: \emph{Is this output correct?} For agents, the evaluation must ask: \emph{Did the system behave correctly over time, under uncertainty, given its constraints?} What matters is consistency, failure modes, and sensitivity to change. As a result, evaluation cannot be reduced anymore to checking final outputs against a reference label. This shift has several consequences. First, variability becomes a signal, not noise. An agent that succeeds once but fails intermittently may be unacceptable, even if its average score looks strong. Second, evaluation must account for environment setup, tool contracts, and hidden assumptions. A failure may reflect an ambiguous task or a broken harness rather than a capability gap. Third, evaluation becomes inseparable from engineering. Prompt formats, inference settings, and execution context are part of what is being measured.

\section{Evaluation as a Core System Function}
\label{sec:core-function}

Evaluation enables controlled iteration and should be understood as a core development function rather than a downstream verification step. AI agents do not produce single outputs but trajectories composed of multiple decisions, tool invocations, and state changes. The consensus among practitioners is that evaluations are the \emph{real moat}. While models, prompts, and data are commoditized, the ability to systematically measure performance is the differentiator between a demo and a product. Moving from subjective ``vibes'' to objective evidence requires an iterative, disciplined methodology.

In practice, the absence of systematic evaluation leads teams to rely on reactive signals such as user reports or production incidents, resulting in brittle systems and slow, risk-averse progress. By contrast, robust evaluation transforms system modifications into testable experiments, allowing improvements to accumulate safely over time. As Anthropic has observed, evaluation capacity increasingly functions as a strategic asset: teams with strong evaluation practices can iterate rapidly while maintaining control, whereas teams without them face high-friction upgrades, prolonged manual testing, or deployment paralysis~\cite{anthropic2026evals}.

Evaluation must instead operate as a continuous process that assesses behavior over time, across variations in inputs, environments, and execution contexts. In this sense, evaluation serves as the human calibration layer: the control mechanism that enables humans to place justified trust in AI systems as operational tools. By translating human intent and risk constraints into observable signals, evaluation becomes the mechanism through which oversight is exercised in practice. It is important to distinguish evaluation from AI risk assessment, as the two are frequently conflated despite serving fundamentally different functions. Risk assessment is prospective and structural: it identifies potential harms, their severity, likelihood, and time to impact, and delineates what is acceptable or unacceptable for an organization. Evaluation, by contrast, is empirical and execution-level; it tests whether failures actually occur, how frequently, under which conditions, and whether system changes increase or reduce exposure.

In short, risk assessment defines expectations and the space of concern and risk, while evaluation enforces those expectations in practice. Without evaluation, risk remains largely declarative and symbolic. Conversely, without risk framing, evaluation collapses into an isolated exercise devoid of normative grounding. The two are therefore tightly coupled and mutually dependent. Evaluation ultimately functions as an enforcement mechanism that operationalizes governance by translating normative constraints into measurable conditions and linking them to concrete decisions such as deployment, rollback, or escalation. In this sense, evaluation is not merely diagnostic but decisional (Annex~I).

Evaluation results directly condition organizational decisions such as deployment approval, scope expansion, rollback, or the granting of autonomy. As such, evaluation outputs must be auditable. Evaluation suites, thresholds, and grading logic constitute part of the system's compliance evidence, supporting internal audit, external assurance, and post-incident analysis. In the absence of traceable evaluation artifacts, accountability degrades into informal judgment and retrospective justification.

Based on insights from industry best practices~\cite{husain2026evals, wolfe2024judge, fourrier2025guidebook, schmid2025pass}, effective evaluation follows an 8-step lifecycle and concrete engineering practices, which we have summarized in Annex~II. Evaluation does not need to start large or sophisticated to be effective. In fact, delaying evaluation until systems are ``ready'' is one of the most common failure modes. Early in development, small and imperfect evaluation suites already provide helpful feedback: changes have large effect sizes, failure modes are easier to isolate, and success criteria can still be shaped rather than retrofitted. What matters at this stage is not coverage, but signal. As agents become more capable and autonomous, these early evaluation artifacts can be progressively hardened into structured suites, reliability checks, and regression guards. Evaluation capacity, therefore, grows incrementally with the system, rather than appearing fully formed at deployment. A pragmatic roadmap for building evaluation capacity from the earliest stages to trusted systems is outlined in Annex~III~\cite{anthropic2026evals, wolfe2024judge, langchain2025deepagents}.

\subsection{Evaluation as a Regulatory Requirement Without a Technical Definition}
\label{sec:regulatory}

Regulators increasingly rely on evaluation as an operational control mechanism, yet they remain notably restrained in specifying how evaluation should be conducted in practice. Across jurisdictions, regulatory texts consistently acknowledge the necessity of evaluation while stopping short of defining concrete benchmarks, metrics, or methodologies.

The European Union's AI Act (Regulation (EU) 2024/1689) anchors evaluation as a regulatory requirement while remaining deliberately non-prescriptive. Providers of high-risk AI systems are required to operate a quality management system (Article~17), which implicitly mandates institutionalized and repeatable evaluation practices. For general-purpose AI model providers, Article~55 goes further by explicitly requiring them to ``perform model evaluation in accordance with standardised protocols and tools reflecting the state of the art.'' Yet beyond this formulation, the regulation offers little guidance on what such an evaluation should concretely entail. The AI Act thus formalizes evaluation as a legal obligation without stabilizing its technical meaning, effectively shifting the burden of definition and implementation onto providers at a time when evaluation practices themselves remain contested and fast-evolving.

In the United States, this pattern is even more pronounced. In the absence of a comprehensive federal AI statute, evaluation is articulated primarily through standard-setting rather than binding law. NIST's AI Risk Management Framework (AI RMF 1.0) formalizes Test, Evaluation, Verification, and Validation (TEVV) as a core lifecycle activity. Under the ``Measure'' function, the RMF calls for documented test sets, metrics, and evaluation tools (Measure~2.1), treating evaluation artifacts as inputs to governance and accountability decisions. However, it explicitly avoids prescribing what constitutes sufficient evaluation. Sectoral regulators adopt a similar stance. The U.S. Food and Drug Administration's guidance on the use of AI in regulatory decision-making introduces a ``credibility assessment framework'' tied to the context of use, framing evaluation as evidentiary support for regulatory claims rather than as compliance with fixed performance thresholds. Here again, regulators specify \emph{why} evaluation is required, but not \emph{how} it should be operationalized.

Taken together, these regimes reveal a common regulatory posture: evaluation is treated as indispensable, yet deliberately under-specified. Regulators define evaluation as an evidentiary and control function---gating deployment, conditioning modification, and supporting oversight---while deferring technical detail to industry practice. This restraint reflects an acknowledgment that evaluation is deeply technical, fast-evolving, and tightly coupled to system architecture, making prescriptive regulation both fragile and quickly obsolete. Even if evaluation is an engineering concern, ownership cannot sit solely with engineering teams. In agentic systems, evaluation artifacts increasingly act as governance instruments: they encode risk thresholds, acceptable behaviors, escalation triggers, and conditions for autonomy. Evaluation must therefore be formally integrated into broader AI governance frameworks, with clear ownership, documentation, and accountability, so that technical performance and robustness signals inform organizational decision-making, compliance, and oversight.

\section{A Taxonomy of Evaluation Objectives}
\label{sec:taxonomy}

The fundamental question driving evaluation is deceptively simple: \emph{How do we know if a model is good?} However, as we move from partially-static Large Language Models (LLMs) to compound AI systems and agents, the definition of ``good'' has fragmented. Evaluation is now the discipline of metrology applied to non-deterministic software~\cite{welty2019metrology} rather than a single step in the classical machine learning pipeline. In addition, evaluation is not a single activity, but a set of distinct practices serving different objectives. Additionally, evaluation is not designed for a single audience, a single system, or a use case. It serves different groups with different incentives. Many of the current confusions around benchmarks, leaderboards, and ``capabilities'' stem from treating these objectives as interchangeable.

To reason clearly about evaluation, it is therefore necessary to separate three motivations that are often conflated in practice:

\begin{enumerate}
    \item \textbf{Non-regression testing (the builder's view):} This is analogous to unit testing in software engineering. When we alter a training recipe, data mixture, or architectural parameter, we must ensure the model has not ``broken''. Here, the absolute score matters less than the trajectory. We look for stability or improvement within expected ranges. If a 7B parameter model suddenly drops 20 points on MMLU\footnote{Massive Multitask Language Understanding.} after a data refresh, the specific score is irrelevant---the signal of regression is what matters, as noted in Hugging Face's analysis~\cite{welty2019metrology}.

    \item \textbf{Leaderboards and rankings (the user's view):} This is the comparative analysis required for model selection. Whether examining the Open LLM Leaderboard\footnote{The Open LLM Leaderboard compares models using standardized benchmark tests.} or the LMSYS Chatbot Arena\footnote{The LMSYS Chatbot Arena compares models by showing their answers to people and asking which one is better.}, the goal is to rank architectures to identify the best proxy for a specific capability. However, rankings are snapshots in time. A leaderboard score is only valid insofar as the underlying benchmark remains unsaturated and uncontaminated~\cite{fourrier2025guidebook}. The ``lessons learned'' from ImageNet suggest that while individual scores fluctuate, stable rankings across broad groups of evaluations provide the only robust signal~\cite{fourrier2024llmeval}.

    \item \textbf{Scientific capability assessment:} This category asks ``Where does the field stand?'' and attempts to measure abstract constructs such as reasoning or world knowledge. It is the most methodologically fragile form of evaluation, as there are no rigorous or widely accepted frameworks for defining these capabilities even in humans, let alone in probabilistic models.
\end{enumerate}

Historically, we relied on static benchmarks such as WordNet (1985) and Modified National Institute of Standards and Technology (1994). In the Language Models era, we moved to General Language Understanding Evaluation (GLUE), SuperGLUE, and eventually MMLU~\cite{fanelli2023benchmarks101}. But as we will discuss, the lifecycle of a benchmark---from ``impossible'' to ``saturated''---is accelerating. MMLU, once a gold standard for professional knowledge, is now effectively solved by frontier models, pushing us toward harder, dynamic, and agentic evaluations~\cite{fanelli2024benchmarks201}.

\subsection{How Do We Frame an Evaluation Pipeline?}
\label{sec:pipeline}

An evaluation pipeline is not merely a dataset; it is an instrument. To construct one, we must rigorously define the inputs, the inference method, and the grading logic.

Before a model produces a single token, the evaluation pipeline already shapes the outcome. Tokenization choices such as how numbers are segmented, how whitespace is handled, or how non-English text is compressed can substantially affect measured performance. Models trained with specific chat templates, system prompts, or role markers may fail dramatically when evaluated with raw text or mismatched prompt formats. Reproducibility, therefore, depends on precise alignment between the model's expected input distribution and the benchmark's prompt structure. Even seemingly minor details, such as token spacing or the handling of start-of-sentence tokens in models like Gemma\footnote{Gemma refers to a family of open-weight large language models released by Google.}, can lead to large performance shifts~\cite{fourrier2025guidebook}.

\paragraph{Agent addendum.} For agentic systems, the notion of ``input'' extends beyond tokens. It includes the task specification, tool interfaces, and the initial environment state (e.g., filesystem structure, caches, or mock services). When the task specification is underspecified---for example, when required file paths are omitted but later assumed by the grader---the evaluation ceases to measure capability and instead reflects implicit guesswork or evaluator assumptions~\cite{anthropic2026evals}.

\subsubsection{The Three Pillars of Evaluation Methodologies}
\label{sec:three-pillars}

The current literature identifies three primary methodological approaches to model evaluation, each characterized by distinct trade-offs in scalability, fidelity, and epistemic reliability~\cite{fourrier2024llmeval}.

\paragraph{Pillar 1---Automated benchmarking.}
Automated benchmarks rely on programmatic scoring without human intervention. Two dominant paradigms are commonly employed.

\begin{itemize}
    \item \textbf{Log-likelihood-based evaluation} is widely used in multiple-choice settings such as MMLU. Rather than generating an answer, the model is evaluated on the conditional probability it assigns to predefined candidate tokens (e.g., ``A'', ``B'') given the input context. This approach is computationally efficient, deterministic, and easy to reproduce. However, it may overestimate performance by favoring models that are well calibrated to the benchmark's prompt structure or answer format, rather than those that exhibit robust task-level reasoning.

    \item \textbf{Generative evaluation} requires the model to produce free-form outputs. This paradigm is necessary for tasks such as code generation, long-form question answering, and agentic behavior. Its primary challenge lies in scoring: valid outputs may be diverse, syntactically dissimilar, and semantically equivalent, making reliable automated grading difficult.
\end{itemize}

\paragraph{Pillar 2---Human evaluation.}
Human-based evaluation remains a central reference point and is typically implemented at varying levels of rigor.

\begin{itemize}
    \item \textbf{Informal testing (``vibe checks'')} consists of ad hoc, developer-driven probing of model behavior. While methodologically weak, such checks often serve as an early detection mechanism for obvious failures.

    \item \textbf{Arena-style evaluations}, such as those based on Elo-style pairwise comparisons (e.g., LMSYS), rely on crowd-sourced human preferences between competing model outputs. These methods capture perceived quality and usability but are inherently sociotechnical: they reflect human preferences, which may systematically favor traits such as verbosity or assertiveness rather than factual accuracy or reasoning depth.

    \item \textbf{Expert annotation} represents the highest-quality form of human evaluation, particularly for domain-specific tasks. However, its high cost, limited throughput, and latency make it unsuitable for continuous evaluation pipelines. In practice, expert judgments are primarily used to calibrate or validate more scalable metrics.
\end{itemize}

\paragraph{Pillar 3---Model-as-a-judge evaluation.}
Recent advances in frontier models (e.g., GPT-4, Claude, Gemini) have enabled the use of large language models as evaluators of other models, serving as a proxy of human evaluation. This approach can be implemented either pairwise (preference judgments) or pointwise (scalar ratings). While highly scalable and well-suited to generative tasks, this methodology introduces a recursive dependency: evaluations inherit the biases, priors, and failure modes of the judge model itself. As a result, the validity of the evaluation is conditioned on the alignment and robustness of the judging model~\cite{wolfe2024judge}.

\subsection{Challenges of Creating Benchmarks, Metrics, and Pipelines}
\label{sec:challenges}

Designing an evaluation framework is often more difficult than training the model itself. As noted by Hamel Husain\footnote{Hamel Husain is a machine learning practitioner and researcher whose work focuses on evaluation, error analysis, and production ML systems, with extensive applied experience in industry.}, teams frequently drown in ``prefab'' metrics and automated tools while skipping the most vital step: looking at the data. The friction points in evaluation range from the philosophical (what is ``natural''?) to the mechanical (tokenizer bugs).

\subsubsection{Challenge 1: The Trap of ``Naturalness'' and the Usefulness Task}
\label{sec:challenge1}

A benchmark is only as good as its alignment with reality. Ofir Press\footnote{Ofir Press is a machine learning researcher and practitioner whose work focuses on evaluation, benchmarking, and the limitations of standard metrics in modern ML systems.} identifies a critical failure mode: evaluating models on tasks that humans never perform~\cite{press2024benchmarks}. Many benchmarks resemble ``IQ tests''---artificial pattern-matching in diagrams or multiple-choice logic puzzles---rather than natural tasks. The ``Usefulness Criteria'' asks: Would a system that solves this benchmark actually make a human more productive? For example, SWE-bench succeeds because it is composed of real GitHub issues\footnote{\url{https://github.com/SWE-bench/SWE-bench}}---fixing bugs is a natural, daily task. Conversely, asking a model to classify a sentence's sentiment into 19 distinct categories (a common academic task) is often unnatural and yields low signal~\cite{press2024benchmarks}. If the question setup is unrealistic (e.g., ``Doctor, my elbow hurts, choose option C''), we are optimizing models for synthetic constraints rather than real-world utility~\cite{press2024benchmarks}.

\subsubsection{Challenge 2: The Saturation Velocity and the ``$-200\%$'' Rule}
\label{sec:challenge2}

Benchmarks have a distinct lifecycle: they are born impossible and die saturated. In the early deep learning era, benchmarks like MNIST or ImageNet drove progress for years. Today, the cycle is compressed to months. If a benchmark launches and the top model scores 80\%, the benchmark is already dead~\cite{press2024benchmarks}. It offers no headroom for improvement. A robust modern benchmark should target a starting accuracy of 0.1\% to 10\% for frontier models. Ofir Press, creator of SWE-Bench, even suggests a ``$-200\%$'' mindset: create questions so hard that even if models improve 3x, they would still score zero today. The idea from Ofir Press is to design benchmarks so hard that today's models all fail, ensuring the benchmark remains meaningful even as models rapidly improve. This future-proofs the instrument against the rapid ascent of model capabilities~\cite{press2024benchmarks}.

\subsubsection{Challenge 3: The ``Error Analysis'' Gap and Generic Metrics}
\label{sec:challenge3}

There is a prevailing myth that evaluation can be fully automated with ``off-the-shelf'' metrics. But, usually error analysis---manually reviewing traces to build a taxonomy of failure modes---is the single most important activity in evals~\cite{husain2026evals}.

Metrics such as BERTScore, ROUGE, or ``Helpfulness'' scores are often counterproductive. They measure abstract qualities that may not correlate with user success. A high ROUGE score on a summary does not guarantee factual accuracy; it only guarantees lexical overlap. What are these metrics actually measuring?

\begin{itemize}
    \item \textbf{ROUGE}: counts overlap of words or phrases between an output and a reference text. It answers ``did you use similar wording?'' rather than ``is it correct?''

    \item \textbf{BERTScore}: uses embeddings to judge semantic similarity between an output and a reference. It answers ``do these sentences convey a similar meaning?'' but remains anchored to a reference and does not verify ground-truth correctness.

    \item \textbf{``Helpfulness'' scores} (as commonly used in LLM evaluation): a human/model judges rate whether the response seems helpful, often without verifying correctness or task completion.
\end{itemize}

These metrics optimize for resemblance to a reference, not for user success. They do not ask whether the user obtained the correct answer, made a sound decision, or completed the task. As a result, they systematically reward fluent but wrong behavior. A summary may score highly on ROUGE while introducing an incorrect date or figure; an answer may be rated ``helpful'' because it is confident and well phrased, even when it is unverifiable or false. Overlap-based metrics are also easy to game: longer outputs tend to score better simply by increasing lexical or semantic coverage, even when they add little value or introduce noise.

A similar problem appears in grading scales. Engineers often default to 1--5 Likert ratings for human or model judgments, but these scales are inherently unstable. The distinction between a ``3'' and a ``4'' is subjective, inconsistent across annotators, and prone to central tendency bias. Binary evaluations (Pass/Fail), while coarser, force explicit decisions and consistently produce higher inter-rater reliability, making them better suited for operational evaluation pipelines.

\subsubsection{Challenge 4: The ``Silent Bugs'' of Inference Configuration}
\label{sec:challenge4}

Perhaps the most insidious challenge in evaluation is what we might call ``silent bugs''---inconsistencies in the serving and inference layer that silently skew results without throwing errors. An evaluation score is meaningless without the exact context of how the model was served. This issue manifests through several concrete failure modes at inference and serving time:

\begin{itemize}
    \item \textbf{Chat template trap:} Many modern models are trained with strict chat templates that encode roles such as \emph{system}, \emph{user}, and \emph{assistant}. Passing the system instruction as a plain user prefix, or altering role boundaries, can significantly degrade performance. When the evaluation prompt does not match the model's training format, failures reflect prompt mismatch rather than reasoning limitations.

    \item \textbf{Sampling parameters:} A model evaluated at temperature=0.0 is functionally different from one at temperature=0.1 or with top\_k=50. Seemingly minor defaults in inference libraries (like vLLM vs.\ Hugging Face \texttt{generate()}\footnote{Hugging Face \texttt{generate()} is a method inside a model library, often used in research and prototyping. vLLM is a dedicated high-performance inference engine optimized for serving and batching at scale.}) can lead to non-reproducible variance.

    \item \textbf{Hardware indeterminism:} Model outputs are not perfectly invariant to hardware and numerical precision. Running the same evaluation on different accelerators (e.g., A100 vs.\ H100\footnote{A100 and H100 are two generations of NVIDIA AI chips. The H100 is newer and faster, and it performs some calculations differently.}) or with different precisions (FP16 vs.\ BF16\footnote{Two ways to represent numbers with limited precision to accelerate AI models.}) can change token generation paths, especially in long or borderline decisions. Without controlling for these factors, evaluation results may not be comparable across runs.

    \item \textbf{Hidden truncation:} ``Silent'' context window truncation occurs when a benchmark prompt exceeds the model's limit, but the inference server simply drops the end of the prompt without warning as a fallback mechanism. The model then fails the task, not because of a lack of capability, but because it never saw the question~\cite{fourrier2025guidebook}. These failures appear as capability gaps but are in fact artifacts of the serving layer.
\end{itemize}

For agents, silent bugs extend beyond inference into the harness and environment~\cite{anthropic2026evals, langchain2025deepagents}. Shared state between runs (leftover files, cached data, contaminated databases) can create correlated failures that look like ``capability regressions''. Shared state can also inflate performance (e.g., agents benefiting from artifacts produced in prior trials). And, resource exhaustion (disk, memory, rate limits) can create non-independent trials, making score deltas meaningless.

A robust agent evaluation must start each trial from a clean slate and treat the harness as part of the measurement instrument, not ``just plumbing''~\cite{anthropic2026evals}.

\subsubsection{Challenge 5: Data Contamination and Goodhart's Law}
\label{sec:challenge5}

The ``dirty secret'' of LLM evaluations is contamination. Models are trained on the internet, and benchmarks are published on the internet. Models often memorize the test set rather than learning the skill~\cite{reiter2024challenges}. The example of the canary's failure is particularly interesting here. As highlighted in \emph{Latent Space}\footnote{Latent Space is an AI newsletter and blog that analyzes model capabilities, benchmarks, and evaluation issues.}, when GPT-4 results were released, BigBench was excluded because the model had memorized the ``Canary GUID'' string---a unique identifier meant to signal ``do not train on this text''~\cite{fanelli2023benchmarks101}. The canary failure illustrates that contamination is not only a data hygiene problem but also a measurement failure. Once benchmarks become optimization targets, models are incentivized to exploit them rather than learn the underlying capability. This dynamic is a direct instance of Goodhart's law: ``\emph{When a measure becomes a target, it ceases to be a good measure}'' said Charles Goodhart\footnote{Charles Goodhart is a British economist.}. The obsession with Needle-in-a-Haystack (NIAH) performance has led models to over-attend to irrelevant details, potentially degrading their performance on summarization tasks where ignoring noise is essential~\cite{yan2024appliedllms}.

\subsubsection{Challenge 6: The Biases of LLM-as-a-Judge}
\label{sec:challenge6}

When using a model to evaluate another, we import specific, documented cognitive biases~\cite{wolfe2024judge, hf2024evalguide}:

\begin{itemize}
    \item \textbf{Position bias:} Models overwhelmingly prefer the first option presented in a pairwise comparison. Mitigating this requires swapping orders and averaging results (position switching).

    \item \textbf{Verbosity bias:} Judges tend to rate longer answers higher, regardless of quality. Techniques like ``Length-Controlled AlpacaEval'' attempt to regress this out, but it remains a persistent issue.

    \item \textbf{Self-Preference bias:} Models tend to rate outputs from their own model family higher.

    \item \textbf{Sycophancy:} Models (and human judges) often prefer answers that confirm their existing biases or are stated assertively, even if factually wrong.
\end{itemize}

Beyond bias, agent-grade judges often show inconsistent strictness and overconfidence when evaluation rubrics are underspecified. In practice, judge prompts often need iteration and calibration against human experts before they can be trusted~\cite{anthropic2026evals, wolfe2024judge}. A simple reliability trick is to give the judge an ``escape hatch'' (e.g., allow ``Unknown'' when evidence is insufficient) to reduce hallucinated verdicts~\cite{anthropic2026evals}.

\subsubsection{Challenge 7: Reliability vs.\ Capability}
\label{sec:challenge7}

Recent work on ``Platinum Benchmarks'' suggests that many standard benchmarks are riddled with label errors. If an LLM gets a question ``wrong'' because the ground truth is incorrect (a pervasive issue in crowd-sourced datasets), correctness is penalized~\cite{vendrow2025reliability}. Furthermore, benchmarks rarely test reliability. A model might solve a hard math problem once but fail it five times if the temperature is slightly perturbed~\cite{vendrow2025reliability}. For agents, low scores can be an evaluation bug rather than a capability signal. Agent evaluations have subtle failure modes that can depress scores even when the agent is strong: ambiguous task specifications, brittle grading, stochastic tasks that cannot be exactly reproduced, or harness constraints that block valid strategies. When you see patterns like ``0\% pass across many trials'' treat them as red flags for broken tasks/graders before concluding the model is incapable.

Here are the most common ``silent'' causes of misleading agent scores~\cite{anthropic2026evals}:

\begin{enumerate}
    \item Starting too late: waiting until production makes it harder to formalize success criteria.
    \item Unrepresentative task sets: evals that don't reflect real user failures (support queue, bug tracker).
    \item Ambiguous task specs: missing constraints/paths/format requirements lead to noisy or unfair failures.
    \item Broken tasks / misconfigured graders: the reference solution doesn't pass; graders assume unstated details; numeric/format brittleness.
    \item One-sided problem sets: only testing ``should do X'' incentivizes over-triggering (e.g., searching everywhere).
    \item Harness and environment flakiness: shared state, caches, infrastructure noise, or resource exhaustion corrupt the independence of trials.
    \item Overly rigid grading of the path: punishing correct outcomes because tool call sequences differ.
    \item No partial credit: masking ``mostly right'' improvements in multi-step workflows.
    \item LLM judge unreliability: bias + hallucinated certainty without calibration.
    \item Bypassability: agents ``cheat'' via loopholes rather than solving the intended problem.
    \item Transcript blindness: not reading traces makes it impossible to distinguish agent mistakes from eval mistakes.
    \item Saturation blindness: 100\% pass gives a regression signal but no improvement signal.
    \item No ownership: suites decay without maintenance, review, and contribution workflows.
\end{enumerate}

\section{Models vs.\ Agents}
\label{sec:models-vs-agents}

As we transition from analyzing models (stateless text-in/text-out predictors) to evaluating agents (autonomous systems that maintain state, use tools, and execute loops), our evaluation frameworks must undergo a fundamental architectural shift. The complexity moves from the prompt to the harness.

\subsection{Anatomy of an Agent Evaluation}
\label{sec:anatomy}

Evaluating an agent is not merely about checking an answer key; it is about auditing a workflow. Anthropic defines the structure of an agent evaluation through four key components~\cite{anthropic2026evals}:

\begin{enumerate}
    \item \textbf{The Harness (Scaffold):} The infrastructure that allows the model to act as an agent (processing inputs, orchestrating tool loops). Crucially, the harness itself is part of what is being tested. A restrictive harness can artificially cap intelligence, while a loose one can introduce noise. For example, Opus~4.5 solved a flight-booking task by finding a loophole in the policy---it ``failed'' the specific constraints of the eval but successfully solved the user's problem.

    \item \textbf{The Transcript (Trace):} The complete record of the trial, encompassing not just the final output, but the reasoning steps, tool calls, and intermediate environmental states. Reading transcripts is essential to distinguish between genuine mistakes and grading bugs.

    \item \textbf{The Outcome:} The final state of the environment: ``Did the file get edited?'', ``Was the calendar invite sent?'' This is distinct from the model's text response ``I have booked the flight''.

    \item \textbf{The Graders:} The logic that scores the performance.
\end{enumerate}

\subsection{The ``Grader'' Ecosystem}
\label{sec:graders}

For agents, we cannot rely on a single score. We must employ a hybrid grading strategy that mirrors the complexity of the agent's tasks~\cite{anthropic2026evals}.

\begin{itemize}
    \item \textbf{Strategy 1---Code-based graders (deterministic):} These are fast, cheap, and objective. They include string matching, static analysis (linting), and state verification (e.g., checking if a specific entry exists in a SQL database). They are excellent for verifying outcomes but brittle for assessing behavior.

    \item \textbf{Strategy 2---Model-based graders (probabilistic):} These are necessary for nuance. Using LLM-as-a-Judge with specific rubrics allows to assess qualitative dimensions, such as the politeness of a conversational agent or the comprehensive nature of a research summary.

    \item \textbf{Strategy 3---Human graders:} While unscalable, they remain the gold standard for calibrating the model-based graders to ensure the LLM judges haven't drifted from ground truth.
\end{itemize}

\subsection{The Reliability Metric: Pass@k vs.\ Pass\texorpdfstring{$^k$}{k}}
\label{sec:passk}

In the world of agents, consistency is the primary bottleneck. Traditional benchmarks often mask reliability concerns by focusing on optimistic scenarios. We need to distinguish between capability and reliability~\cite{schmid2025pass}:

\begin{itemize}
    \item \textbf{Pass@k} measures capability: it captures the probability that at least one correct solution appears within $k$ attempts. This metric is appropriate in settings such as code generation or ideation, where multiple outputs can be sampled, and a human can select the best one.

    \item By contrast, \textbf{Pass$^k$} measures reliability: it represents the probability that all $k$ trials succeed without failure. Formally, it is defined as:
    \begin{equation}
        \text{Pass}^k = \left(\frac{c}{n}\right)^k
    \end{equation}
\end{itemize}

If an agent has a 70\% success rate on a single flight rebooking task, Pass@3 might look impressive at ${\sim}97\%$ (chance of at least one success). However, Pass$^3$ reveals a disaster: the chance of handling three consecutive requests without failure is only 34.3\%~\cite{schmid2025pass}. For autonomous agents operating without human-in-the-loop, Pass$^k$ is the true measure of production readiness.

\subsection{The Lifecycle: Capability to Regression}
\label{sec:lifecycle}

The goal of an evaluation suite changes over time. Initially, we build capability evaluations---hard tasks with low pass rates (e.g., ${<}10\%$) to drive hill-climbing and measure breakthrough intelligence. As the agent matures, these morph into regression evaluations---tasks that must maintain near 100\% pass rates. When an eval becomes ``saturated'' (solved), it doesn't become useless; it becomes a guardrail against backsliding~\cite{anthropic2026evals}.

\subsection{Cost and Efficiency Frontiers}
\label{sec:cost}

Finally, agentic evaluation must account for the Pareto frontier of performance vs.\ compute. OpenAI's analysis on \emph{BrowseComp} and \emph{PaperBench}\footnote{Agent evaluation benchmarks introduced by OpenAI.} highlights that performance often scales linearly with test-time compute. Therefore, a ``better'' agent isn't just one with higher accuracy; it is one that achieves that accuracy with fewer tokens, fewer tool calls, and lower latency. We must penalize agents that brute-force solutions via infinite loops, even if they eventually arrive at the correct state~\cite{openai2025browsecomp, openai2025paperbench}.

\section{Evaluations with Environments}
\label{sec:environments}

The ultimate frontier in evaluation is the environment. As models graduate from chatbots to assistants, they no longer operate in the vacuum of a prompt; they operate in stateful, noise-filled, and dynamic worlds. As explained above, the previous generation of ``read-only'' benchmarks (like MMLU) is insufficient for agents that must read, write, and adapt. Evaluation instruments should shift from static datasets to simulated environments.

\subsection{GAIA2: The ``Messy Reality'' of AI Assistants}
\label{sec:gaia2}

If GAIA (2023) was about retrieval, GAIA2 (2025) is about \emph{agency in a chaotic world}~\cite{fourrier2025gaia2}. Released by Meta and Hugging Face, GAIA2 represents a paradigm shift from simple question-answering to managing the complexity of a ``read-and-write'' existence. GAIA2 runs within the Agent Research Environment (ARE), a simulated smartphone ecosystem that includes functional applications: Email, Calendar, Contacts, and FileSystem. Unlike static evaluations, the agent doesn't just answer a question; it changes the state of the world (e.g., ``Add the Renne family to the birthday invite on Nov 7th''). GAIA2 explicitly tests capabilities that break traditional models.

To reflect real-world conditions, GAIA2 explicitly tests capabilities that are poorly captured by traditional benchmarks. It evaluates whether an agent can handle ambiguity by requesting clarification instead of hallucinating a solution when instructions conflict. It introduces controlled tool failures, such as timeouts or API errors, to test whether the agent can recover gracefully rather than failing outright. It also stresses temporal reasoning, a known weakness of current frontier models---because their internal world model often lacks a coherent clock---by requiring agents to manage time-dependent actions such as delayed scheduling or sequencing tasks relative to future events (e.g., ``Book a cab \emph{after} a 3-minute delay'').

GAIA2 moves beyond raw accuracy. A model that solves a task in 3 minutes with 500 tokens is superior to one that takes 30 minutes and 50k tokens. Evaluation must now plot agents on a Pareto frontier of score vs.\ monetary cost, penalizing inefficiency even in successful traces~\cite{fourrier2025gaia2}.

\subsection{TextQuests: Long-Context Reasoning in Simulated Worlds}
\label{sec:textquests}

While GAIA2 tests practical utility, TextQuests uses the constrained, adversarial worlds of interactive fiction (specifically 25 titles developed by Infocom such as Zork), where agents must navigate, explore, and solve problems purely through language to test intrinsic reasoning and exploration~\cite{phan2025textquests}. Text adventure games are particularly well suited to stress-testing reasoning under long contexts. The environment unfolds step by step, causing the context window to grow continuously, often beyond 100k tokens. To succeed, an agent must maintain a persistent ``mental map'' of the world over thousands of steps without external tools.

TextQuests exposes several failure modes that are largely invisible in static benchmarks.

\begin{itemize}
    \item \textbf{Spatial disorientation:} Models often fail to reverse a sequence of actions. If an agent climbs up a cliff, it frequently cannot figure out how to climb down, indicating a failure to model the reversibility of physical space.

    \item \textbf{State hallucination:} As context grows, models suffer from ``inventory amnesia'', confidently claiming to hold an item (e.g., a matchbook) that they dropped hundreds of turns ago.

    \item \textbf{Dynamic thinking \& efficiency:} TextQuests highlights a massive inefficiency in current agent architectures. Agents treat every step---whether it's a trivial ``move north'' or a complex puzzle solution---with the same computational weight. A truly effective agent must demonstrate dynamic thinking, allocating reasoning budget (e.g., Chain-of-Thought) only when the complexity of the state demands it~\cite{phan2025textquests}.
\end{itemize}

\subsection{The Agentic Evaluation Loop}
\label{sec:eval-loop}

Evaluating in these environments requires a fundamentally different architecture than the simple ``Prompt $\rightarrow$ Response'' loop of 2023. We now utilize ``Harnesses'' or ``Sandboxes'' (like ARE, Harbor, or custom Docker containers) to ensure reproducibility in non-deterministic settings~\cite{anthropic2026evals, fourrier2025guidebook, langchain2025deepagents}:

\begin{itemize}
    \item \textbf{Setup (Clean State):} Deep agents are stateful. To avoid ``contamination'' between test cases, the environment must be wiped and re-seeded (e.g., a fresh file system or database) for every single trial~\cite{langchain2025deepagents}.

    \item \textbf{Execution (The loop):} The harness runs the agent loop: \emph{Observation $\rightarrow$ Thought $\rightarrow$ Action $\rightarrow$ Observation}. This allows us to capture the full trace, not just the final output.

    \item \textbf{Teardown \& Verification:} We grade the side effects. We do not ask the LLM if it sent the email; we check the mock SMTP\footnote{SMTP (Simple Mail Transfer Protocol) is the standard protocol used to send emails.} server logs. This moves grading from subjective text analysis to deterministic state verification.

    \item \textbf{Extensibility (MCP):} Modern environments like ARE are designed to be extensible via the Model Context Protocol (MCP). This allows researchers to plug in new tools (e.g., a robot arm interface or a live API) to test agents on entirely new modalities without rewriting the evaluation harness~\cite{fourrier2025gaia2}.
\end{itemize}

\section{Conclusion}
\label{sec:conclusion}

The central challenge ahead is not that AI systems are becoming more capable, but that our ability to measure and constrain their behavior is failing to keep pace. Evaluation has quietly shifted from a supporting role to a defining one: it now determines which systems can be trusted, which failures are rendered visible, and which forms of progress are even legible to organizations.

As agentic systems proliferate, evaluation evolves into a dual discipline. It is a discipline of measurement, tasked with making behavior observable, comparable, and auditable over time. But it is also increasingly a form of moderation, in the sense that evaluation now gates autonomy, bounds acceptable behavior, and triggers escalation, rollback, or human intervention.

The field, therefore, faces a choice. We can continue to optimize for scores that are easy to produce yet weakly grounded in real-world behavior, or we can treat evaluation as a first-class measurement and control function---one that demands rigor, maintenance, and explicit assumptions. The latter path is slower, less glamorous, and harder to automate, but it is the only one compatible with deploying autonomous systems in high-stakes environments.

In this sense, the future of AI progress may depend less on architectural breakthroughs than on whether we can build evaluation instruments that keep pace with the systems they are meant to govern. Aligned with the motivation behind the Open Benchmarks Grants, recently announced by Snorkel AI and partners where they argued as well that we are currently trapped in an ``evaluation gap''~\cite{chen2026evalgap} where our ability to develop agents has outpaced our ability to measure them. Closing this gap is the only way to convert the current hesitation around high-stakes deployment into justified trust. However, closing this gap cannot be exclusively an Anglocentric endeavor. We urgently need similar grant initiatives dedicated to multilingual benchmarks and under-served languages and regions, preventing a future where agentic autonomy is robust in English but fragile everywhere else. What matters is not whether agents and/or models can succeed once, but whether failure becomes rare, interpretable, and bounded. The open question for the field is no longer how intelligent our systems can become, but how much uncertainty we are willing to tolerate---and how precisely we can measure and moderate it.

\bibliographystyle{unsrt}
\bibliography{references}

\clearpage
\appendix

\section*{Annex I --- Risk Assessment and Evaluation as Part of AI Governance}
\addcontentsline{toc}{section}{Annex I}
\label{annex:governance}

\begin{figure}[h!]
    \centering
    \includegraphics[width=\textwidth,height=0.85\textheight,keepaspectratio]{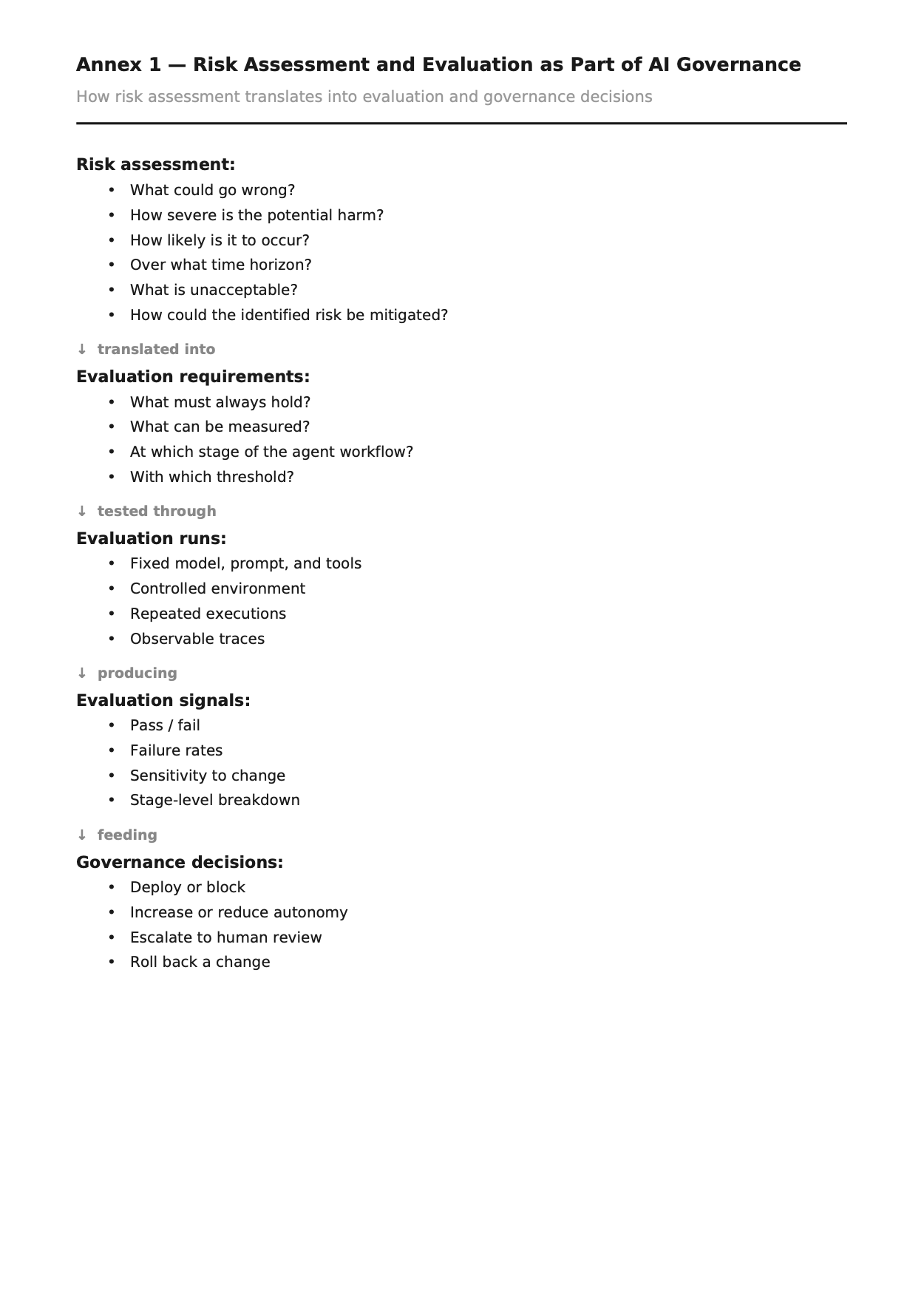}
    \caption{How risk assessment translates into evaluation and governance decisions.}
    \label{fig:annex1}
\end{figure}

\clearpage
\section*{Annex II --- AI Evaluation Best Practices}
\addcontentsline{toc}{section}{Annex II}
\label{annex:best-practices}

\begin{figure}[h!]
    \centering
    \includegraphics[width=\textwidth,height=0.85\textheight,keepaspectratio]{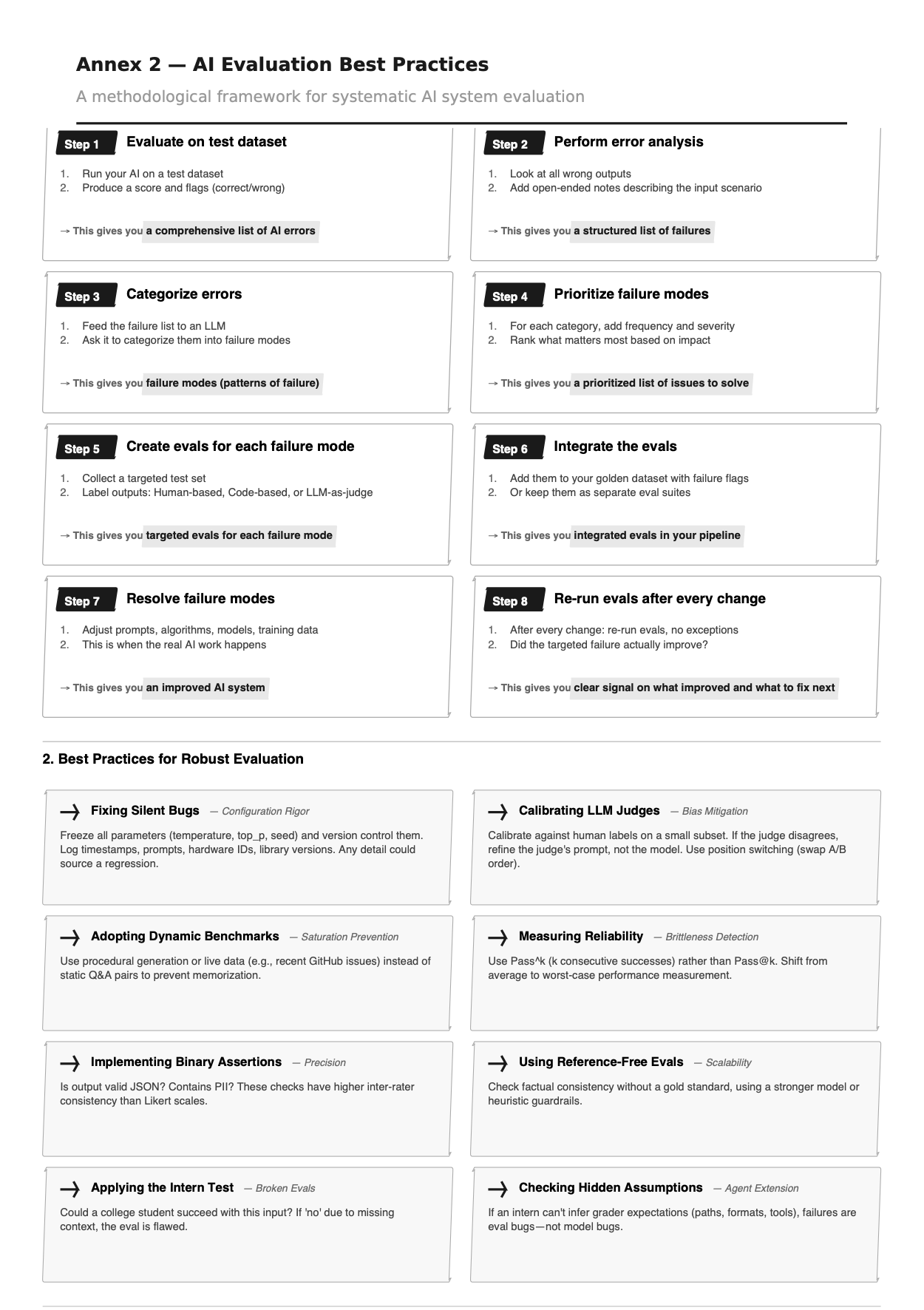}
    \caption{A methodological framework for systematic AI system evaluation.}
    \label{fig:annex2}
\end{figure}

\clearpage
\section*{Annex III --- A Roadmap from ``No Evals'' to ``Trusted Evals''}
\addcontentsline{toc}{section}{Annex III}
\label{annex:roadmap}

\begin{figure}[h!]
    \centering
    \includegraphics[width=\textwidth,height=0.85\textheight,keepaspectratio]{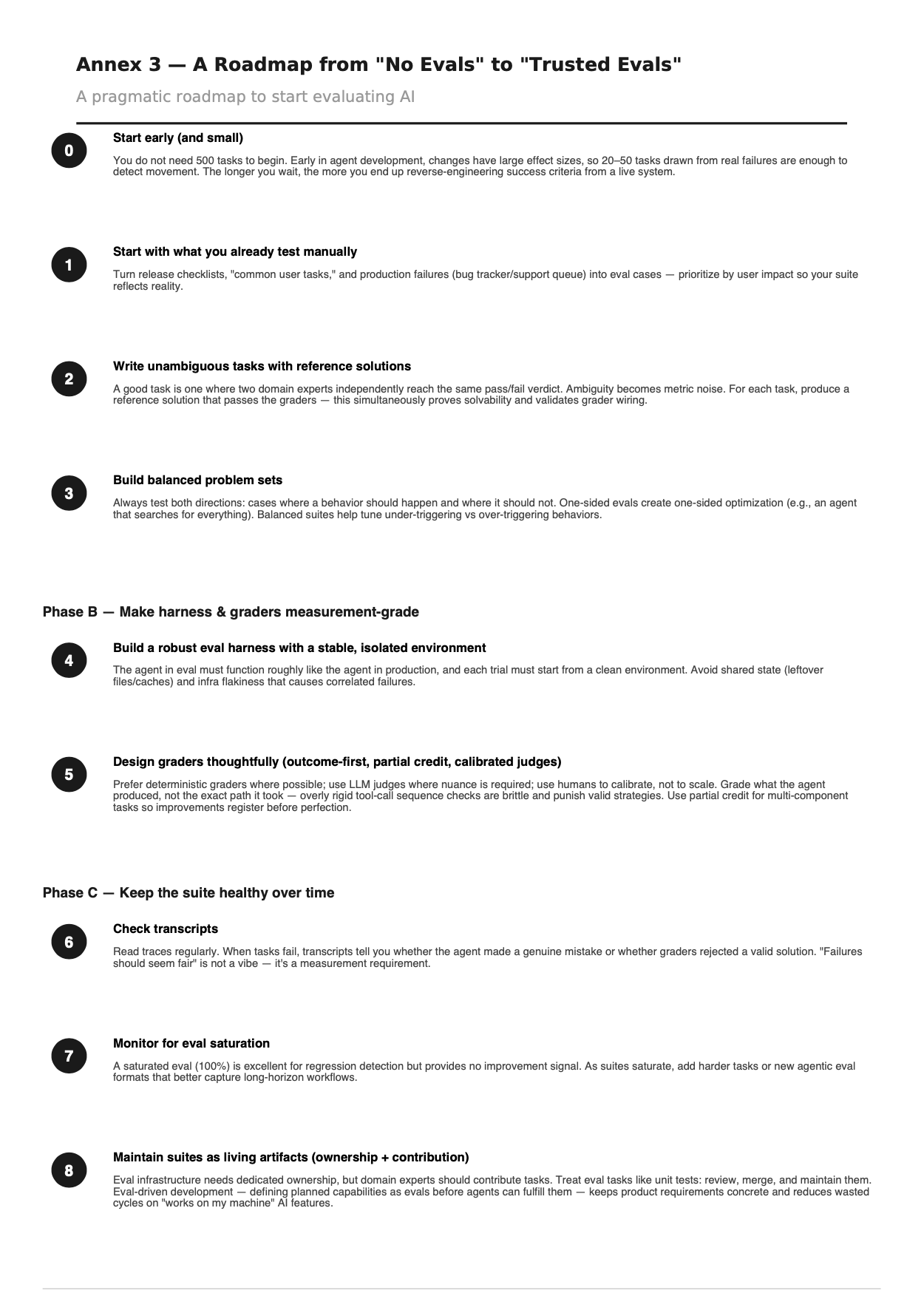}
    \caption{A pragmatic roadmap to start evaluating AI.}
    \label{fig:annex3}
\end{figure}

\end{document}